\documentclass[preprint,12pt,number]{elsarticle}

\usepackage[T1]{fontenc} 

\usepackage{multicol}
\usepackage{comment}

\usepackage{tabularx} 

\usepackage{amsmath} %
\usepackage{amssymb} %
\usepackage{amsfonts} %

\DeclareMathOperator*{\conc}{\bigoplus}

\journal{Computers in Biology and Medicine}

\begin{document}

\begin{frontmatter}



\title{Transformation trees - documentation of multimodal image registration}


\author[label1]{Agnieszka Anna Tomaka\corref{cor1}}
\author[label1]{Dariusz Pojda}
\author[label1]{Michał Tarnawski}
\author[label1]{Leszek Luchowski}
\affiliation[label1]{organization={Institute of Theoretical and Applied Informatics, Polish Academy of Sciences},
            addressline={Baltycka 5}, 
            city={Gliwice},
            postcode={44-100}, 
            country={Poland}}

\cortext[cor1]{atomaka@iitis.pl}

\begin{abstract}
Multimodal image registration plays a key role in creating digital patient models by combining data from different imaging techniques into a single coordinate system. This process often involves multiple sequential and interconnected transformations, which must be well-documented to ensure transparency and reproducibility. In this paper, we propose the use of transformation trees as a method for structured recording and management of these transformations. This approach has been implemented in the dpVision software and uses a dedicated .dpw file format to store hierarchical relationships between images, transformations, and motion data.
Transformation trees allow precise tracking of all image processing steps, reduce the need to store multiple copies of the same data, and enable the indirect registration of images that do not share common reference points. This improves the reproducibility of the analyses and facilitates later processing and integration of images from different sources. The practical application of this method is demonstrated with examples from orthodontics, including the integration of 3D face scans, intraoral scans, and CBCT images, as well as the documentation of mandibular motion.
Beyond orthodontics, this method can be applied in other fields that require systematic management of image registration processes, such as maxillofacial surgery, oncology, and biomechanical analysis. Maintaining long-term data consistency is essential for both scientific research and clinical practice. It enables easier comparison of results in longitudinal studies, improves retrospective analysis, and supports the development of artificial intelligence algorithms by providing standardized and well-documented datasets. The proposed approach enhances data organization, allows for efficient analysis, and facilitates the reuse of information in future studies and diagnostic procedures.
\end{abstract}


\begin{highlights}
\item A tool for documentation of multimodal image registration 
\item Simplification of new image  registration
\item Storing an image processing history
\item The dpVision tool supports the transformation tree method
\end{highlights}


\end{frontmatter}



\section{Introduction}

The development of computer systems and tools for acquiring multidimensional, multimodal images has led to the increasing presence of the terms \textit{virtual patient} and \textit{virtual twin} in the literature \cite{joda2014,marradi,seth}. Both concepts refer to the acquisition of digital patient data and their alignment in a way that provides the most complete and accurate representation of the patient, both in terms of their current condition and changes occurring over time. These approaches are particularly useful in orthodontics and for supporting surgical planning. The \textit{virtual patient} allows surgeons to thoroughly analyse a patient’s anatomy in advance, anticipate potential difficulties, and virtually simulate a procedure based on a personalized 3D digital model. Similarly, the \textit{virtual twin} enables monitoring of changes in the patient's condition over time, supporting clinical decision-making and allowing rapid intervention in case of complications. 

The creation of a digital patient representation consists of several key steps: image acquisition, segmentation and 3D reconstruction, multimodal data integration, and physical or biomechanical simulations. This workflow is widely applied not only in orthodontics but also in other areas of medicine \cite{LO2023100615}. However, to fully exploit the potential of these technologies, accurate registration and synchronization of data from different imaging modalities are essential.

Data registration is the process of aligning objects from different sources to a common reference frame. It is widely used in various fields, including scientific visualization, satellite image analysis, computer graphics, and augmented reality. The integration of multimodal imaging data requires transforming individual images so they can be visualized in a shared patient-related coordinate system, enabling precise analysis and comparison. 

The registration process involves determining a transformation that minimizes the differences between the compared datasets. Depending on the type of data, different techniques are applied:

\begin{itemize}
    \item \textbf{Rigid registration} --- assumes no changes in shape and allows only translation and rotation.
    \item \textbf{Affine registration} --- additionally permits scaling and shearing.
    \item \textbf{Non-rigid registration} --- allows local deformations and is commonly used in biological image analysis.
\end{itemize}

Image registration requires calibration of imaging systems to ensure proper scaling and metric data acquisition. It can be based on known geometric relationships between coordinate systems or determined directly from the images. If corresponding points are known, alignment can be performed using the \textit{Least Squares} method \cite{Golub1973LeastSquares}. When correspondences are unknown, the \textit{Iterative Closest Point (ICP)} algorithm \cite{besl} estimates them iteratively while simultaneously determining the transformation. In cases where images lack common structures, additional reference objects can be used to maintain spatial consistency.

Registration of moving objects requires distinguishing between stable and dynamic regions, aligning the stable regions to analyse motion. Different types of transformations are applied depending on the image characteristics: rigid for fixed structures, affine for CT scans with gantry tilt, non-rigid for flexible anatomical structures, and projection transformations for aligning 3D images with 2D projections.

Transformations can be applied as a single operation or as a sequence of intermediate registrations. However, multiple transformations may introduce numerical inaccuracies. These inaccuracies are particularly problematic in medical applications, where precision and data integrity are critical. Therefore, it is essential not only to justify each registration step but also to document the entire process properly.

This paper presents a method for documenting the image registration process, addressing many challenges found in commonly used solutions. Some of these issues, particularly in the medical field, are discussed in Section~\ref{sec:motivation}, which outlines the motivation behind this work.

Section~\ref{sec:method} presents our original model for representing relationships between images using transformation trees, along with the method for its implementation. Subsection~\ref{subsec:tree-concept} presents some of the mathematical issues necessary to introduce the concept of a transformation tree. Subsection~\ref{subsec:tree-implementation} explains how transformation trees operate in the \textbf{dpVision} software \cite{POJDA2025102093,dpvision}, an advanced tool developed at IITiS PAS for processing multimodal images. Finally, Subsection~\ref{subsec:dpw-format} is dedicated to the \texttt{.dpw} file format, specifically designed to store transformation trees along with rich metadata, ensuring seamless integration into image analysis workflows.

Section~\ref{sec:example} illustrates how the proposed method can be applied in practice, with a particular focus on orthodontic applications. It demonstrates the critical role of multimodal registration in integrating CBCT data, 3D facial scans, and intraoral scans. However, it is important to note that precise image registration and its documentation are also essential in other medical fields. Transformation trees can be useful in maxillofacial surgery, oncology, and biomechanical analysis.

Finally, Sections~\ref{sec:discussion} and \ref{sec:conclusions} provide a discussion and conclusions on the proposed method.

\section{Motivation}\label{sec:motivation}

The use of multimodal imaging in medicine requires precise registration and synchronization of images from different sources so that they can be analysed within a common patient-related coordinate system. This enables accurate analysis and comparison of data from various imaging techniques, such as \textit{CBCT}, \textit{3D surface scans}, and \textit{intraoral scans}. A key element of this process is image registration, which ensures alignment with the actual anatomical structure of the patient, regardless of the imaging modality or original coordinate system. This is achieved by determining a transformation that minimizes the distance between corresponding points, allowing the fusion of images from different sources into a single reference system. However, effective registration involves more than just aligning images --- it also requires proper documentation to ensure reproducibility and integration with other data. The lack of systematic documentation can lead to the loss of crucial information, errors in data analysis, and limited possibilities for reusing registration results.

\subsection{Challenges in Data Documentation}

While many tools enable effective image registration, significantly less attention is given to the systematic documentation of this process. Managing and recording image transformations is a major challenge, especially when multiple sequential transformations are applied. Each image is initially acquired in its own coordinate system, and the transformation process may involve both rigid and affine transformations. When transformations are applied iteratively or at different time points, errors may accumulate, leading to a loss of data integrity. 

Another difficulty is the lack of a standardized method for storing transformations. Many registration tools do not document transformations in a way that allows their reuse --- some provide only the transformed images, while others return a transformation matrix. However, few approaches systematically record the entire process. If the user does not manually save this information, transformations that exist only temporarily on the screen are lost once the program is closed, making it impossible to perform subsequent analysis, reprocessing, or validation of the registration results.

Another problem is data redundancy. To prevent information loss, some systems store multiple copies of the same images in different coordinate systems. This leads to inefficient memory management and complicates further data operations. Storing numerous transformed versions of images not only increases storage requirements but also makes analysis more difficult, as the user must manually identify which version is relevant. As a result, information becomes fragmented, with no unified structure describing all performed operations.

\subsection{Existing Approaches to Storing Image Transformations}

Current methods for storing registration results vary depending on the system and standard used. Tools such as \textbf{MeVisLab} \cite{Ritter2011MeVisLab}, \textbf{ITK} \cite{McCormick2014ITK}, and \textbf{Elastix} \cite{Klein2010Elastix} provide capabilities for image registration and may return a final transformation matrix or a transformed image. However, they often lack systematic recording of the full registration process. As a result, important intermediate steps are lost, making it difficult to reconstruct the history of transformations or verify results retrospectively. Similar challenges occur in geometry-based registration tools, such as \textbf{RapidForm}, \textbf{MeshLab} \cite{meshlab2008} and \textbf{CloudCompare} \cite{girardeau2014cloudcompare}, which are commonly used for registering surface meshes and point clouds using algorithms like \textit{ICP}. While the data types differ, these tools share the limitation of not storing the full sequence of transformations. Additionally, deformable registration tools such as \textbf{ANTs} \cite{Tustison2021ANTsX}, \textbf{Elastix} \cite{Klein2010Elastix}, and \textbf{3D Slicer} \cite{Fedorov2012Slicer} introduce further complexity in transformation management, yet still provide limited support for comprehensive documentation.

Another challenge is the format used to store registration results. Most programs use their own dedicated formats optimized for specific applications, which makes data exchange between different systems difficult. In addition to software-specific formats, more universal formats such as \texttt{XML} \cite{Hunter2001XML}, \texttt{JSON} \cite{Crockford2006JSON}, and \texttt{glTF} \cite{Khronos2017glTF} are also used, providing greater flexibility for data storage and exchange. Some 3D formats, such as \texttt{OBJ} \cite{Wavefront1992OBJ}, \texttt{STL} \cite{Attene2018STL}, and \texttt{VRML/X3D} \cite{Web3D2010VRML}, have become standard in 3D modeling and printing, but their use for documenting registration transformations is limited.

Some systems store multiple versions of images in different coordinate systems, as seen in programs like \textbf{3DSlicer}. While this approach prevents data loss, it leads to redundancy, increasing memory requirements and complicating image management.

The \textit{DICOM Spatial Registration (DICOM SR)} \cite{Sharp2008DICOM,McCormick2014DICOM} standard allows transformations to be stored as special \texttt{DICOM} objects, representing the most standardized approach. However, its main limitation is that it typically records only the final transformation, without tracking the history of modifications. Additionally, implementing this standard is complex, and not all imaging systems support it. In summary:
\begin{itemize}
    \item Storing only the final image prevents analysis of the registration process.
    \item Storing only the transformation matrix does not preserve the full transformation history.
    \item \textit{DICOM SR} is complex and often stores only the final transformation.
    \item Storing multiple versions of images leads to redundancy and organizational difficulties.
\end{itemize}

Each of the existing methods has its limitations, highlighting the need for a more universal and flexible approach to storing transformations that allows for both their reuse and the reduction of unnecessary data redundancy.

Unlike methods that store only the final transformations or retain redundant copies of images, the approach based on transformation trees enables hierarchical tracking of all applied transformations while minimizing memory usage.

\section{Transformation tree: model, representation, storage format}\label{sec:method}

The \textit{transformation tree} provides a unified mathematical model for documenting and organizing image transformations. Its hierarchical structure allows not only for data integration and accuracy analysis of registration methods but also for full control over transformation quality, which is crucial in scientific and clinical applications. This structure ensures that the complete history of transformations is preserved, facilitating the analysis and validation of results.

In this section, we present the formal model of the transformation tree, its representation and visualization method in the \textbf{dpVision} software, as well as the data storage format in \texttt{.dpw} files.

\subsection{Concept of the transformation tree model}\label{subsec:tree-concept}

Let's define an image representation as a matrix $\mathbf{M} \in \mathbb{R}^{4 \times n}$, which contains homogeneous coordinates of image points or vertices of a mesh or vertices of voxels.
\begin{equation}
\mathbf{M}=\begin{bmatrix}
\begin{array}{ccc}
    x_1 & \hdots & x_n\\
    y_1 & \hdots & y_n\\
    z_1 & \hdots & z_n\\
    1 & \hdots & 1
\end{array}
\end{bmatrix}
\end{equation}

Let the operation
$\conc_{i=1}^{I}\mathbf{M}_{i} =\Big[\mathbf{M}_{1}\quad\hdots\quad\mathbf{M}_{I}\Big]\in \mathbb{R}^{4 \times (n_1+\dots+n_I)}$ be a concatenation of $I$ matrices
 $\mathbf{M}_i \in \mathbb{R}^{4 \times n_i}$ for $i\in 1 \dots I$

As each image or partial model is captured in the coordinate system associated with the device, it has to be transformed to the selected coordinate system during the registration process.

The transformation can be defined as follows:

\begin{equation}
\mathbf{M'}=\mathbf{P}\cdot\mathbf{M}
\label{singletransf}
\end{equation}
where $\mathbf{M}$, $\mathbf{M'}$ are image representations  accordingly in the device and patient-specific coordinate systems; $ \mathbf{P}$ is the transformation matrix.

Let's assume $ \mathbf{P}$ is a linear, reversible transformation. In the case of a rigid body transformation, the matrix $\mathbf{P}$ has the form:
\begin{equation}
 \mathbf{P} = \mathbf{T}\cdot\mathbf{R}\cdot \mathbf{S}\end{equation}
where  $\mathbf{S}$ is the scale, $\mathbf{R}$ is the rotation matrix, $\mathbf{T}$ is the translation matrix.

The  transformation may be represented by a single transformation matrix (equation \ref{singletransf})  or by a sequence of matrices representing partial registrations. For linear transformations, the cumulative transformation matrix is the multiplication of a list of subsequent transformation  matrices of partial scans.
\begin{equation}
\mathbf{M'}=\mathcal{P}\cdot\mathbf{M}
=\prod_{j=J}^{1}{\mathbf{P}_{j}}\cdot \mathbf{M}
= \left(\mathbf{P}_{J}\cdot\ldots\cdot\mathbf{P}_{2}\cdot\mathbf{P}_{1}\right)\cdot\mathbf{M}\end{equation}


where $\mathbf{P}_{J}\cdot...\cdot\mathbf{P}_{2}\cdot\mathbf{P}_{1}$ are the transformation matrices representing subsequent transformations of the model from device to final coordinate systems, $\mathcal{P}$ --- cumulative transformation matrix, $J$ --- the number of subsequent transformations.

An overall model of an object \(\mathcal{M}\)   consists of multiple --- $I$ images or partial models \(\mathbf{M'}_i\) that have been captured in different coordinate systems and then registered in one common model coordinate system using the $\mathcal{P}_{i}$ transformation.

\begin{equation}
    \mathcal{M}=\conc_{i=1}^{I}\mathbf{M'}_{i}=\conc_{i=1}^{I}({\mathcal{P}_{i}\cdot\mathbf{M}_{i}})
    \label{overalmodel}
\end{equation}
where each \(\mathcal{P}_{i}\) transform is a cumulative transformation matrix resulting from $J_{i}$ transformations acting on the $i$-th partial model.



When $K$ partial models are transformed with the same transformation
$\forall k \in {1,...,K}$ $ \mathbf{P}_{k,J_{k}}=\mathbf{P}_{J}$
the equation \ref{overalmodel} can be rewritten:


\begin{equation}\begin{split}
\mathcal{M}&=\conc_{i=1}^{I}{\left(\prod_{j=J_{i}}^{1}{\mathbf{P}_{i,j}}\cdot \mathbf{M}_{i}\right)} = \\
&=\Bigg[\mathbf{P}_{J}\cdot\conc_{i=1}^{K}{\left(\prod_{j=(J_{i}-1)}^{1}{\mathbf{P}_{i,j}}\cdot \mathbf{M}_{i}\right)}\quad\conc_{i=K+1}^{I}{\left(\prod_{j=J_{i}}^{1}{\mathbf{P}_{i,j}}\cdot \mathbf{M}_{i}\right)}\Bigg]
\label{razyplus}
\end{split}\end{equation}

The  concatenation of all elements subject to the same transformation forms a group and it corresponds to the node of the tree. In fact, the overall model also is a group. The transformation between the nodes corresponds to the transformation between the coordinate systems of the nodes.

\begin{figure}[h]
    \centering
    \includegraphics[width=\linewidth]{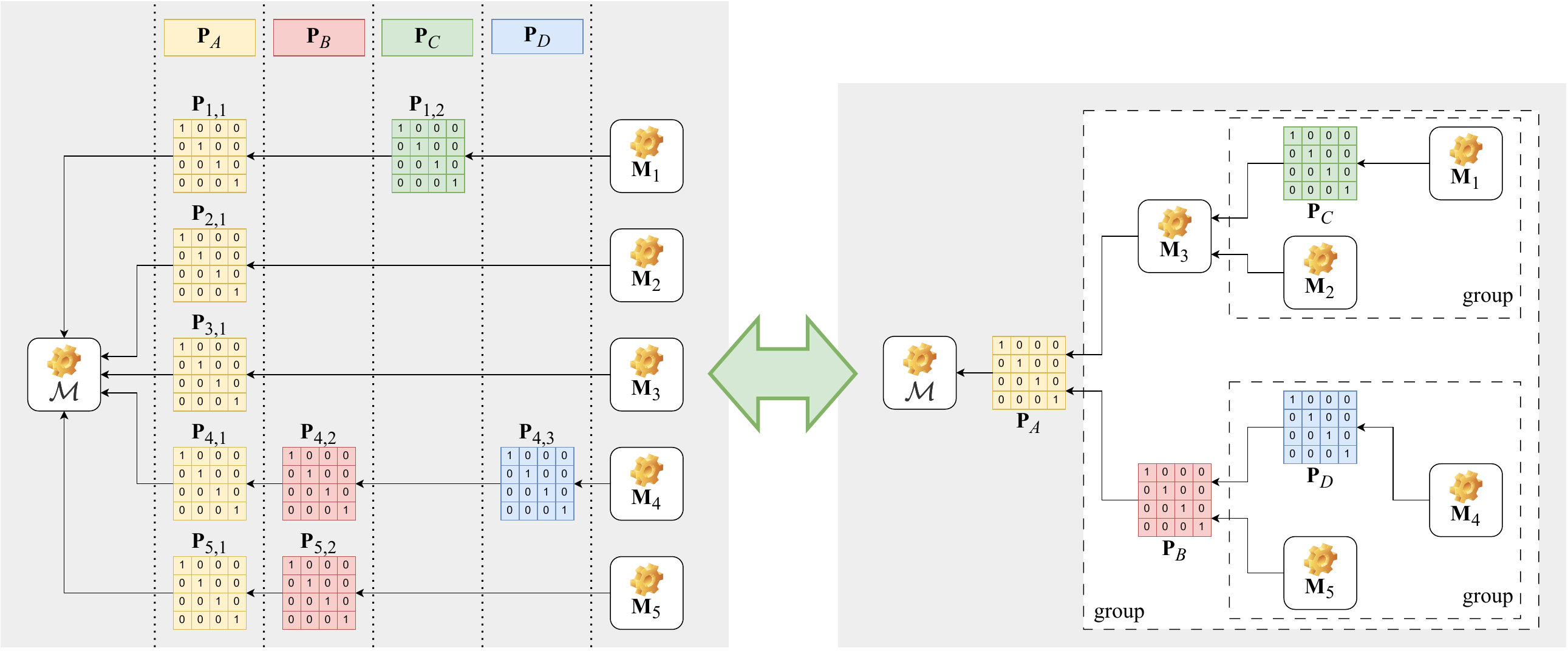}
    \caption{Partial images and lists of transformations for each of them in the example $\mathcal{M}$ model and the equivalent transformation tree, together with selected object groups.}
    \label{fig:sample-tree-model}
\end{figure}

 By recursively finding successive identical transformations and drawing them out before the parenthesis as in the formula (\ref{razyplus}), a tree-like hierarchical structure is obtained, which we refer to as a transformation tree (Figure~\ref{fig:sample-tree-model}).

The application of the mathematical concept of a tree involves establishing relationships between images (partial models) within the model. All data form a unified structure.

It is possible to represent this structure not only in the overall model coordinate system of the model but also in any coordinate system, particularly one associated with any partial model or even in any node (group). When transforming the overall model to the coordinate system of $i$-th partial model, the transformation should be performed using the inverse of the cumulative transformation matrix $\mathcal{P}_{i}^{-1}$.
For transforming the entire model to a given node coordinate system, the inverse of the transformation matrix product from that node to the overall model coordinate system has to be applied.



\begin{figure}[h!]
    \centering
    \includegraphics[width=\linewidth]{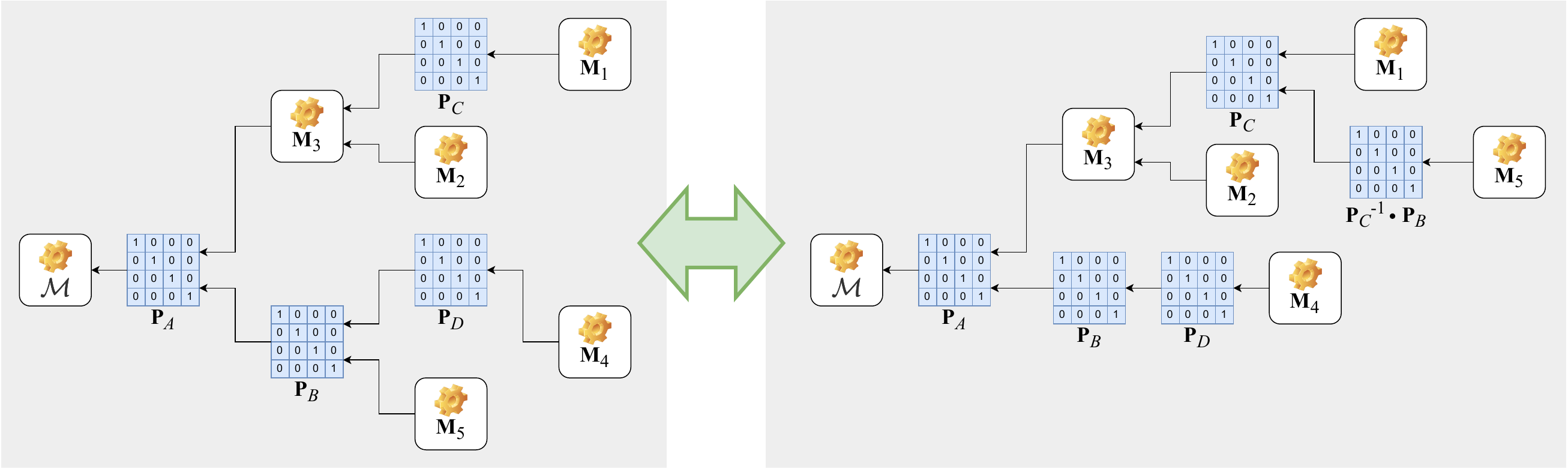}
    \caption{Moving partial images within the transformation tree only requires the use of a single additional transformation matrix to maintain their original position relative to the overall model.}
    \label{fig:moving-in-tree}
\end{figure}

In the mathematical model, the groups have no semantic meaning; furthermore, based on the transformation tree diagram, it is possible to determine how to move an object from one group to another (Figure~\ref{fig:moving-in-tree}). For any couple $a,b$ of partial objects, the equation~\ref{overalmodel} can be transformed into:

\begin{equation}\begin{split}
\mathcal{M}&=\Big[\mathcal{P}_{a}\cdot \mathbf{M}_a \quad \mathcal{P}_{b}\cdot \mathbf{M}_b \quad \conc_{i\neq a,i\neq b}{{\mathcal{P}_{i}}\cdot \mathbf{M}_{i}}\Big]=\\
&=\Big[ \mathcal{P}_{a}\cdot \big[\mathbf{M}_a \quad \mathcal{P}_{a}^{-1}\cdot \mathcal{P}_{b}\cdot\mathbf{M}_b\big] \quad \conc_{i\neq a,i\neq b}{{\mathcal{P}_{i}}\cdot \mathbf{M}_{i}}\Big]
\label{movingobject}
\end{split}\end{equation}

where  $\big[\mathbf{M}_a \quad \mathcal{P}_{a}^{-1}\cdot \mathcal{P}_{b}\cdot\mathbf{M}_b\big]$ forms a new group.

The mathematical concept of the hierarchical transformation tree binds all partial images/models together, allows the partial models to be combined into groups, and provides a diagram of transformation matrices to represent any model or partial model in any coordinate system of any group or model.
The hierarchical structure of the transformation tree makes it possible to trace operations for single images and whole groups of data.

\subsection{Transformation tree representation in dpVision}\label{subsec:tree-implementation}

\textbf{dpVision} is an original software developed at IITiS PAS, designed to handle multimodal data \cite{POJDA2025102093}. Its foundation is the concept of reversible, lossless transformations, which minimize numerical errors and increase processing efficiency.

\begin{figure}[h!]
    \centering
    \includegraphics[width=0.75\linewidth]{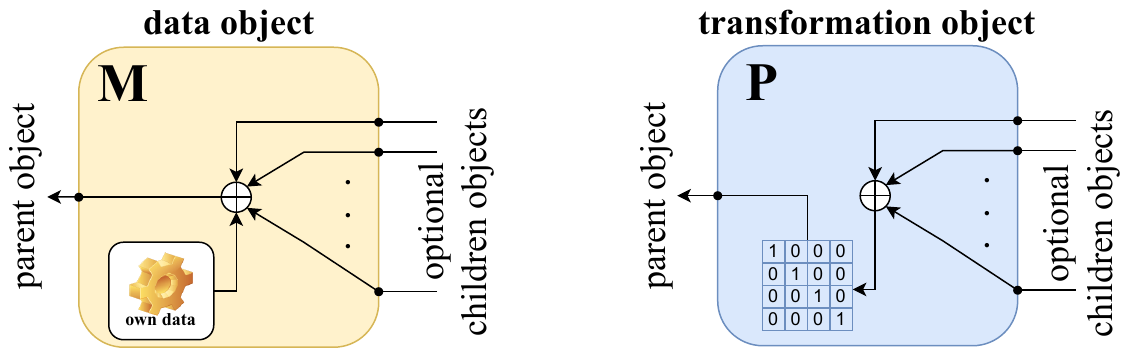}
    \caption{Differences in the interpretation of visual data (left) and transformation (right) as independent objects in \textbf{dpVision}}
    \label{fig:object-def}
\end{figure}

In \textbf{dpVision}, data operations play a key role and, like the data itself, are treated as independent objects. Instead of being assigned to specific scene elements, transformations exist as separate entities, allowing for flexible manipulation and organization within a tree structure. As a result, the software focuses not only on storing data but primarily on processing and dynamically transforming it.

Although the hierarchical \textit{parent} \(\rightarrow\) \textit{child} structure is commonly used in various 3D modeling tools, such as \textbf{Blender} \cite{Hess2010BlenderFoundations,BlenderManual}, or game engines like \textbf{Unity} \cite{Adeniji2024Unity} and \textbf{Unreal Engine} \cite{Kruger2024UnrealEngine}, the representation of transformations in \textbf{dpVision} is unique. In most systems, data operations, especially transformations, are directly assigned to objects as their properties. In \textbf{dpVision}, however, transformations exist as independent objects (Figure~\ref{fig:object-def}), making them an integral part of the scene and simplifying both hierarchy management and editing.

This structure allows for:

\begin{itemize}
    \item \textbf{Hierarchical relationships}: organizing objects in a \textit{parent} \(\rightarrow\) \textit{child} structure, which facilitates scene grouping and manipulation.
    \item \textbf{Transformation sharing}, enabling efficient management of complex datasets.
    \item \textbf{Operation reversibility}, allowing each transformation to be undone without losing the original data.
    \item \textbf{Dynamic coordinate system adjustment}, enabling scene reconfiguration without modifying the original object values.
\end{itemize}

Transformations in \textbf{dpVision} not only define the coordinate system of child objects but also enable flexible scene management. Sequential multiplication of transformation matrices allows for precise control over object positioning, while the ability to rearrange objects within the tree structure without changing their position relative to the overall model (Figure~\ref{fig:moving-in-tree}) or workspace provides users with greater flexibility in scene organization. As a result, the data structure can be reconfigured without the need to recompute original values, significantly improving efficiency when working with large datasets.

\subsubsection{Visualisation of the workspace in the observer's coordinate system}

\begin{figure}[t!]
    \centering
    \includegraphics[width=\linewidth]{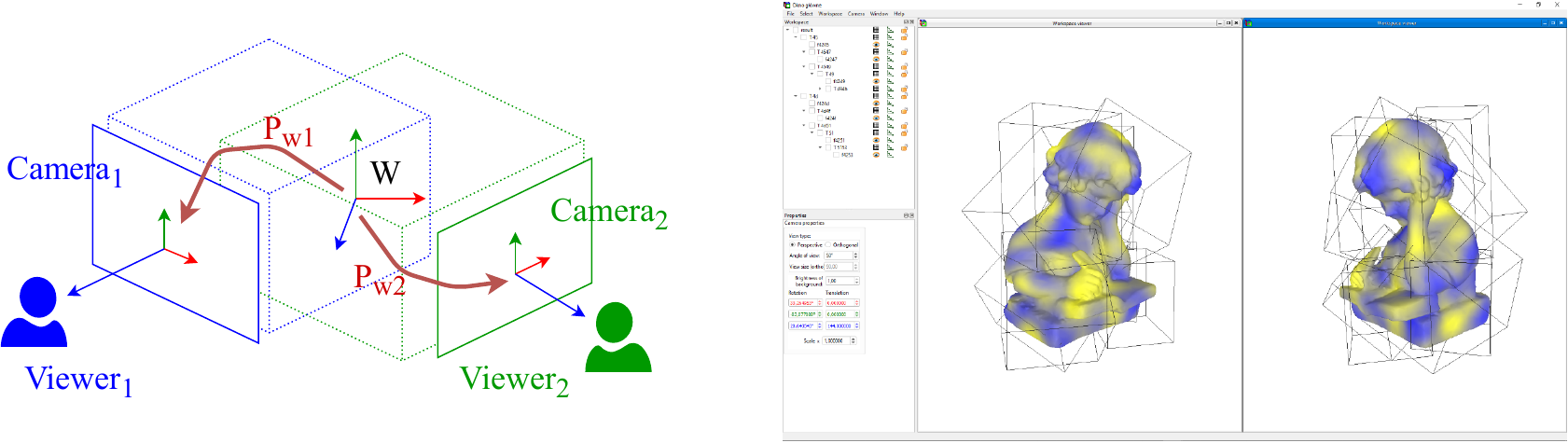}
    \caption{Visualization of the workspace $\mathcal{W}$ in multiple views (schematic – left, \textbf{dpVision} window – right). The transformations $P_{w_1}$ and $P_{w_2}$, representing the camera positions, allow the observer to be placed at different viewpoints.}
    \label{fig:Pw}
\end{figure}

The \textbf{dpVision} software allows for visualization from multiple viewpoints simultaneously, each displayed in a separate window (Figure~\ref{fig:Pw}). The user can freely define the camera position relative to the scene, which in practice requires positioning all displayed models in a coordinate system associated with the selected camera. This is achieved through the camera transformation $\mathbf{P}_w$.  

So far, we have considered the transformation tree model for a single overall model $\mathcal{M}$. However, in practical applications, multiple models are often present in the program's workspace, requiring a consistent way to manage their transformations. Since \textbf{dpVision} supports both multiple viewpoints and multiple models, the workspace can be defined as an $M$-element group of models present in the scene. This group can be expressed as follows:

\begin{equation}
\mathcal{W}=\conc_{m = 1}^{M}{\mathcal{M}_{m}} = \Big[\mathcal{M}_{1}\quad\mathcal{M}_{2}\quad\hdots\quad\mathcal{M}_{M}\Big]
\end{equation}

\noindent
then its representation $\mathcal{W'}$ in the observer's coordinate system is given by:

\begin{equation}
\mathcal{W'}=\mathbf{P}_w\cdot\mathcal{W}=\mathbf{P}_w\cdot\Big[\mathcal{M}_{1}\quad\mathcal{M}_{2}\quad\hdots\quad\mathcal{M}_{m}\Big]
\end{equation}

\noindent
which simplifies to:

\begin{equation}
\mathcal{W'}= \Big[\mathbf{P}_w\cdot\mathcal{M}_{1}\quad\mathbf{P}_w\cdot\mathcal{M}_{2}\quad\hdots\quad\mathbf{P}_w\cdot\mathcal{M}_{m}\Big]
\end{equation}

\subsection{Storage format for .dpw files}\label{subsec:dpw-format}

The \texttt{.dpw} file format has been designed as a text-based, user-readable, and easily editable format for storing scene structures in the form of a transformation tree. Each object in \textbf{dpVision}, whether a visual element or a data operation (e.g., a transformation), can be recorded in a hierarchical structure:

\begin{quote}
\begin{verbatim}
object_type {
    <properties_list>
    <child_objects_list>
}
\end{verbatim}
\end{quote}

\noindent where:
\begin{description}
    \item[\texttt{object\_type}] specifies the type of object (e.g., a triangle mesh, volumetric image, or transformation).
    \item[\texttt{properties\_list}] contains object attributes such as a label, description, references to source files, or a transformation matrix.
    \item[\texttt{child\_objects\_list}] defines child objects, enabling hierarchical structuring.
\end{description}

Geometric data is not stored directly in \texttt{.dpw} files; instead, they contain references to source files and their placement within the scene hierarchy. This approach helps to avoid redundancy and optimizes memory management.

An example of a 3D object stored in \texttt{.dpw} format:
\begin{quote}
\begin{verbatim}
shell {
    label "example_object"
    file "path/to/file.mesh"
    <other_optional_properties>
    <optional_child_objects>
}
\end{verbatim}
\end{quote}

\noindent Similarly, a transformation, instead of a file reference, contains a transformation description in the form of a matrix or its components:
\begin{quote}
\begin{verbatim}
trans {
    label "Identity transformation"
    matrix [ 1.0 0.0 0.0 0.0
             0.0 1.0 0.0 0.0
             0.0 0.0 1.0 0.0
             0.0 0.0 0.0 1.0 ]
    <optional_child_objects>
}
\end{verbatim}
\end{quote}

The hierarchical structure of \texttt{.dpw} files allows for easy scene reconfiguration and enables tracking changes without losing information about the original coordinate system.

\subsection{Summary}

The transformation tree in \textbf{dpVision} provides a consistent approach to image registration and organization. Its hierarchical structure enables tracking the history of transformations, integrating data from different sources, and efficiently managing multimodal image datasets. Additionally, the ability to store transformations in \texttt{.dpw} files facilitates the reconstruction of processing history and the automation of the registration process in image analysis.

Proposed model allows for:
\begin{itemize}
    \item Avoiding data redundancy – storing transformations instead of multiple image versions.
    \item Indirect registration of images without common reference points.
    \item Minimizing registration errors by tracking transformation propagation.
    \item Optimizing the registration process through hierarchical data organization.
\end{itemize}

Hereby, indirect registration is one of the key advantages of the transformation tree.  Such registration refers to the ability to determine the spatial relationship between two images that do not share a direct transformation path. In the transformation tree model, each image is linked to a common root via a unique transformation chain. Therefore, a transformation from image A to image B can be computed by composing the inverse of the path from A to the root with the path from the root to B. This allows images to be registered indirectly, without having to perform direct pairwise registration, as long as both are connected to the tree. This property is especially valuable in clinical applications involving complex, multi-stage acquisition workflows, where certain datasets cannot be registered directly due to missing reference points.

\section{Practical application of the transformation tree: documentation of multimodal image registration in orthodontics.}\label{sec:example}

Traditional orthodontic records typically consist of \textit{3D dental models}, \textit{2D} or \textit{3D X-ray images}, and either \textit{photographs} or \textit{3D scans} of the face, smile, and \textit{intraoral tissues}.

\begin{figure}[h!]
\centering
\includegraphics[width=\linewidth]{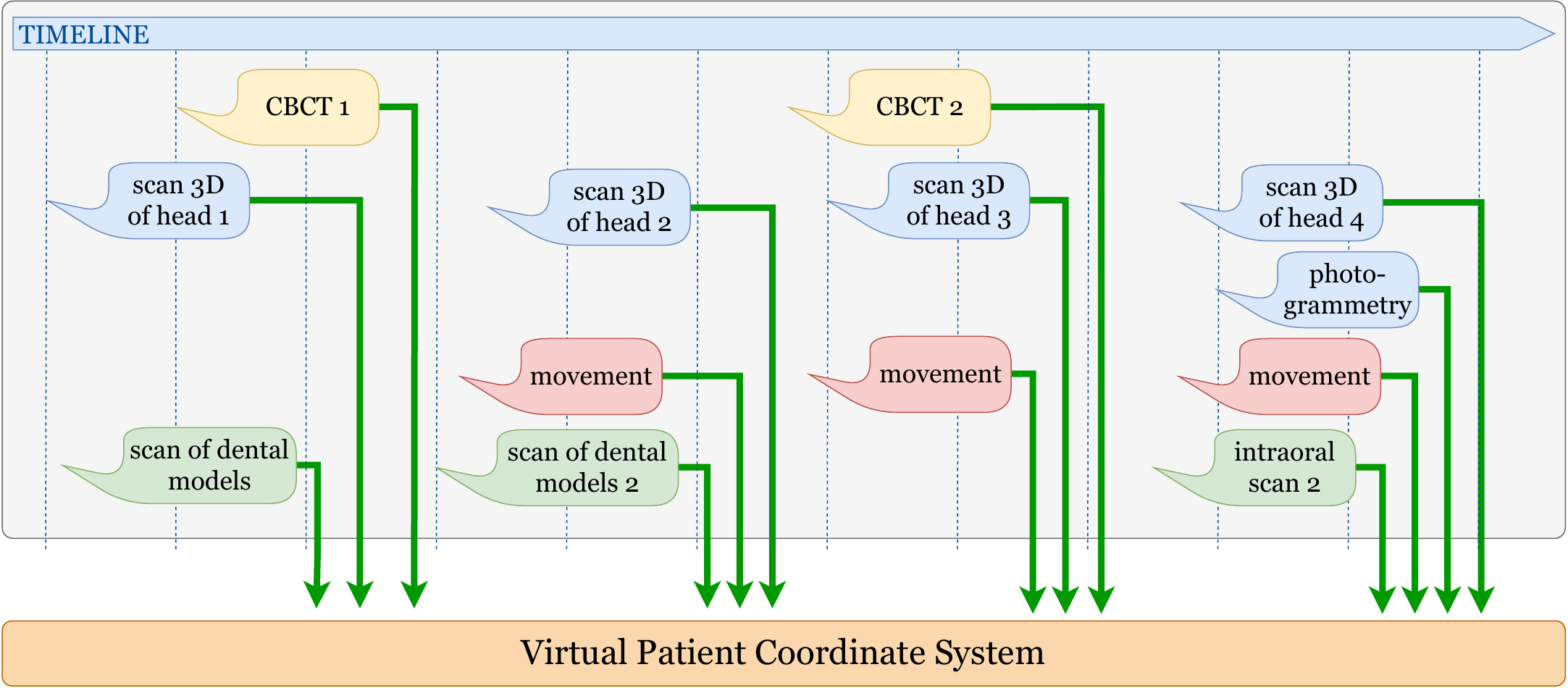}
\caption{Multimodal Patient Data scheme\label{fig1231}}
\end{figure}

This collection of data, diverse in both format and content, already constitutes a multimodal dataset (Figure~\ref{fig1231}).
While each imaging modality can independently contribute to diagnostics, combining them creates a synergistic effect, resulting in a more comprehensive representation, visualization, and understanding of the relevant structures \cite{Bydgoszcz2019}.


\begin{figure}
    \centering
    \includegraphics[width=0.8\linewidth]{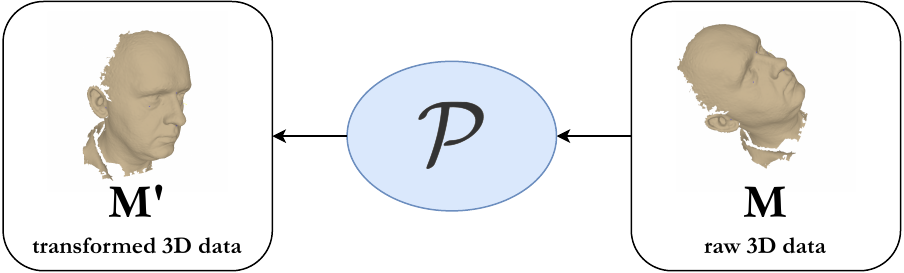}
    \caption{Single transformation scheme (note: in most of figures the convention R to L is used)}
    \label{fig:single_transformation}
\end{figure}

\subsection{Patient-specific coordinate system}

Even a single image is usually transformed to the patient’s coordinate system for visualization and processing (Figure~\ref{fig:single_transformation}). This allows time-based comparisons for the same patient and statistical analysis across patients.

\subsection{ Embeding intraoral scans in 3D face scans --- list of transformations}

The transformation may be represented by a single transformation matrix $ \mathbf{P}$  or by a sequence of matrices representing partial registrations.
At each level, the hierarchical structure is preserved; previous transformations become children of the next one, and the resulting transformation is the multiplication of the matrices from the sequence, up to the root of the tree --- workspace coordinates and down to the device coordinate system.
The simplest example of the sequential use of the transformation matrices is the situation of registration using coarse registration, being the first step before the overall registration. 

\begin{figure}
    \centering
    \includegraphics[width=\linewidth]{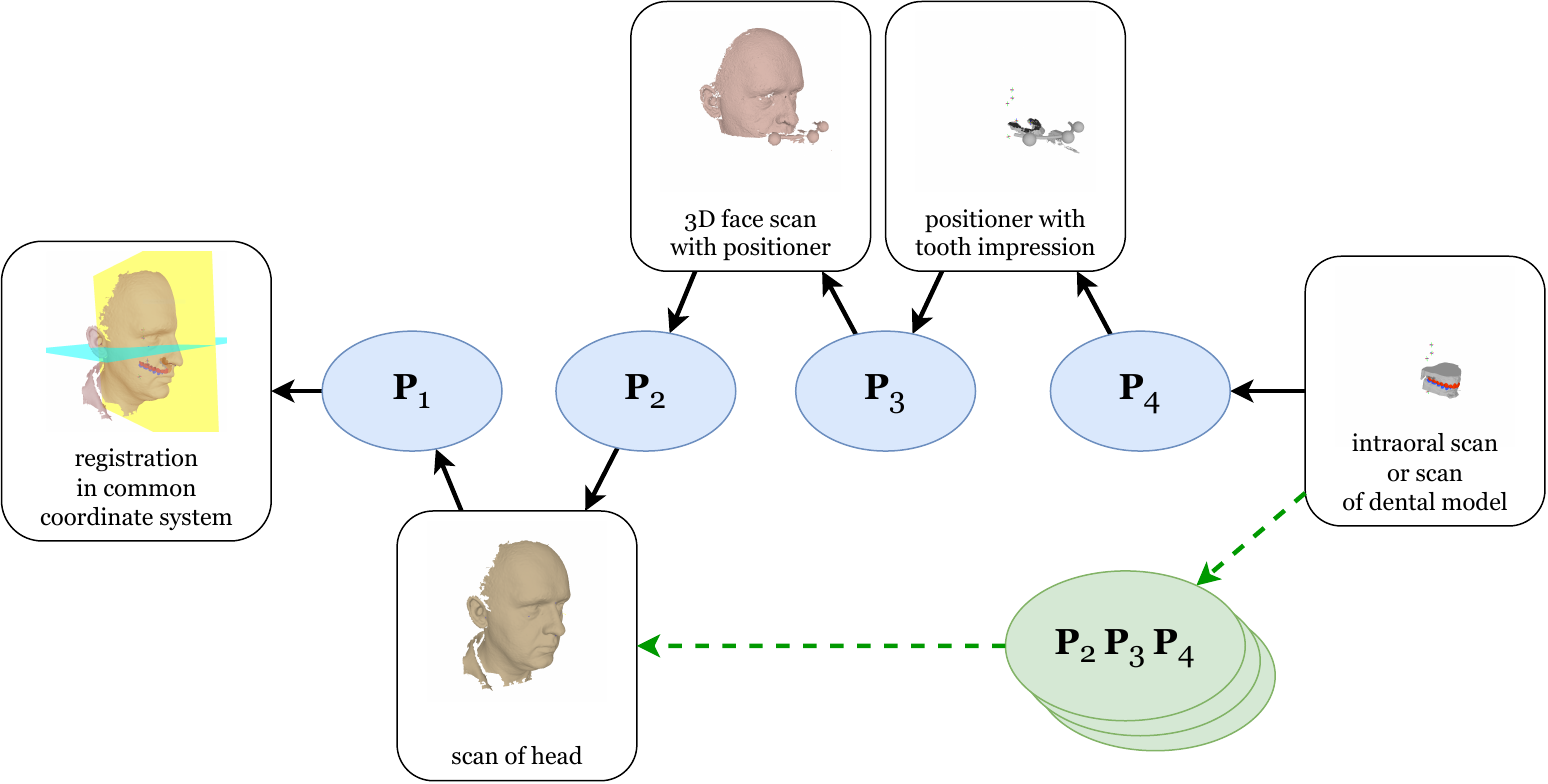}
    \caption{Example of a transformation tree representing the registration of an intraoral scan into a 3D face scan using intermediate scans with a positioner. Each edge corresponds to a transformation matrix ($P_1$ to $P_4$), and the hierarchical structure reflects the sequence of operations. The figure illustrates how multiple partial registrations are combined to form a complete transformation path within a unified coordinate system.}
    \label{fig:transformation_tree}
\end{figure}

Another example that demonstrates the linking of a sequence of transformations is the embedding of intraoral scans inside a 3D face scan. The procedure was presented in \cite{tomaka:2007,solaberrieta2,solaberrieta3,Lama}. 
The idea is to use a facial arch with intraoral and external surfaces for registration. These additional surfaces are scanned with both objects. 
The auxiliary scans make it possible to create a list of transformations required to register 3D face and intraoral scans. 
Adding additional intermediate registrations thus corresponds to nesting the transformations in the list. (Figure~\ref{fig:transformation_tree})  The overall transformation corresponds to the multiplication of the matrices of component transformations of the list.

\subsection{ A group --- common coordinate system }

The tree of transformations allows for the creation of groups of objects undergoing the same transformation. 
A group is a set of images or objects that can be visualized in a common coordinate system. 

In contrast to the mathematical model where semantic meaning was not important, now when creating a group, it is important for the user to know, for example, that the images in question come from the same source. 

\begin{figure}
	\centering
   	\includegraphics[width=\linewidth]{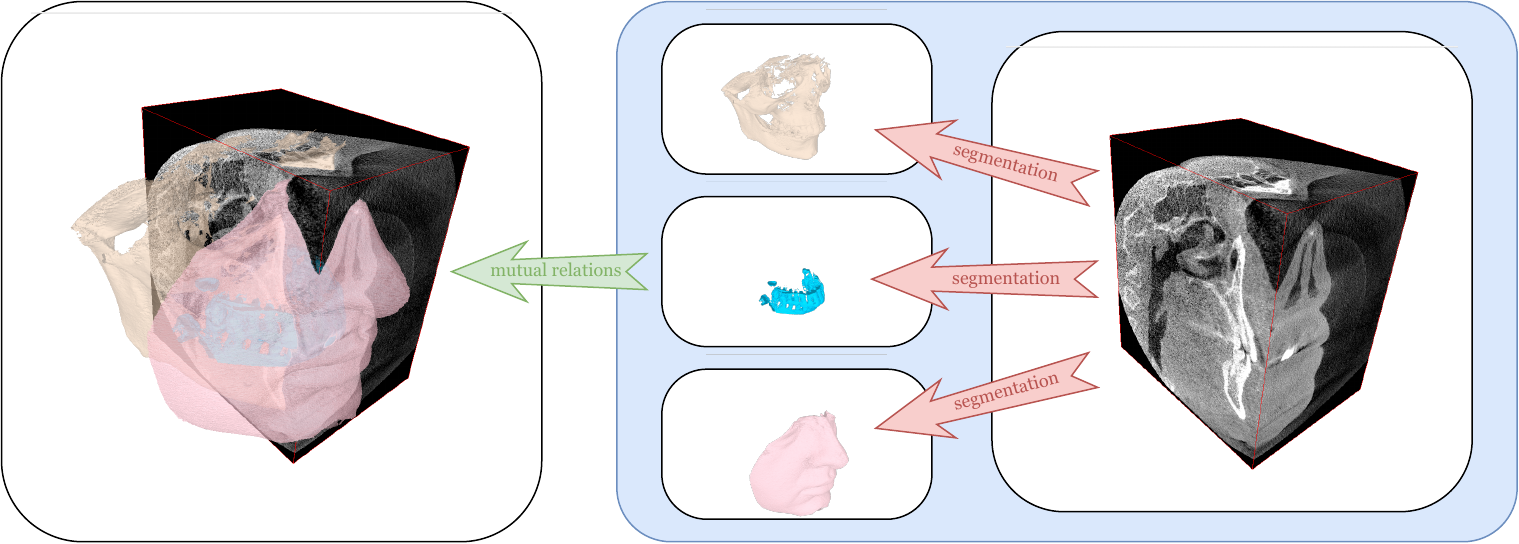}
    \caption{A group formed by CBCT volume data (right) and CBCT segmentation – 3D reconstructions of: bones (top-middle), teeth (center), soft tissues (bottom-middle) and their mutual relations within the group (left)}
    \label{fig:volum}
\end{figure}

Such a group can be formed by segmenting images --- the extracted parts of the image are still in the coordinate system of the image before segmentation. In this case, the volumetric image becomes the parent, and the segmented images are its children.  The cumulative transformation matrix of $\mathcal{P}_{v}$ of the volumetric image becomes the common matrix for all elements of the group.

\begin{equation}
\Big[\mathbf{M'_{v}}\quad\mathbf{M'_{1}}\quad\hdots\quad\mathbf{M'_{i}}\Big] = \mathcal{P}_{v}\cdot
\Big[\mathbf{M_{v}}\quad\mathbf{M_{1}}\quad\hdots\quad\mathbf{M_{i}}\Big]
\end{equation}
where $\Big[\mathbf{M_{v}}\quad\mathbf{M_{1}}\quad\hdots\quad\mathbf{M_{i}}\Big] $ is a concatenation of matrices representing a volumetric image and also different segmentations --- in this case, of teeth, bones, and soft tissues --- of the same volumetric image (Figure~\ref{fig:volum}). In addition, it is possible to separate the bones and the teeth into maxillary and mandibular parts.

\begin{figure}
    \centering
    \includegraphics[width=\linewidth]{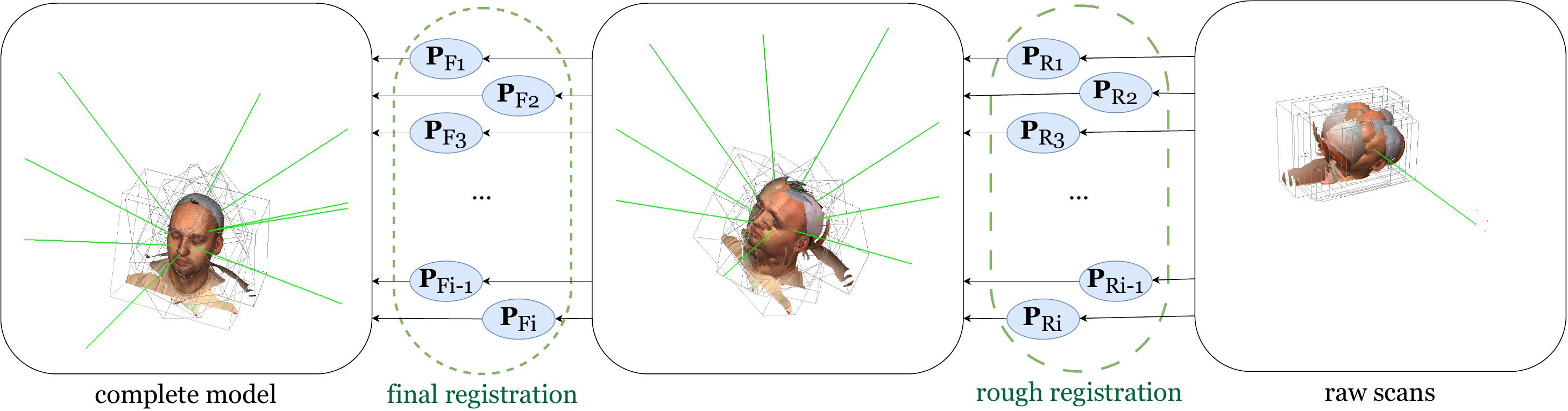}
    \caption{Documentation of registering 3D face scans from different view points. Initial rough registration and final registration}
    \label{fig:heuristicregistering}
\end{figure}

A group can also be formed by elements that have been brought into the same coordinate system as a result of registration.  Figure~\ref{fig:heuristicregistering} shows 3D scans of a face obtained from different viewpoints by a scanner moving around, while the scanner's displacements are known only approximately, but these rough transformations are sufficient to initially position the 3D scans  so that the \textit{ICP} algorithm can find the exact transformations.

\begin{equation}\begin{split}
\Big[\mathbf{M'}_{1}\quad\hdots\quad\mathbf{M'}_{i}\Big] &=
\Big[ \mathcal{P}_{1}\cdot
\mathbf{M}_{1}\quad\hdots\quad\mathcal{P}_{i}\cdot\mathbf{M}_{i}  \Big]=\\
&= \Big[ \mathbf{P}_{F_1}\cdot\mathbf{P}_{R_1}\cdot\mathbf{M}_{1}\quad
 \hdots\quad
 \mathbf{P}_{F_i}\cdot\mathbf{P}_{R_i}\cdot\mathbf{M}_{i}  \Big]
\end{split}\end{equation}


where $\mathbf{P}_{F_i}$ a matrix of a final transformation, $\mathbf{P}_{R_i}$ --- a rough initial transformation of $i$-th partial scan of a head.

\subsection{Groups of objects --- transformation and nesting }

\begin{figure}
    \centering
    \includegraphics[width=\linewidth]{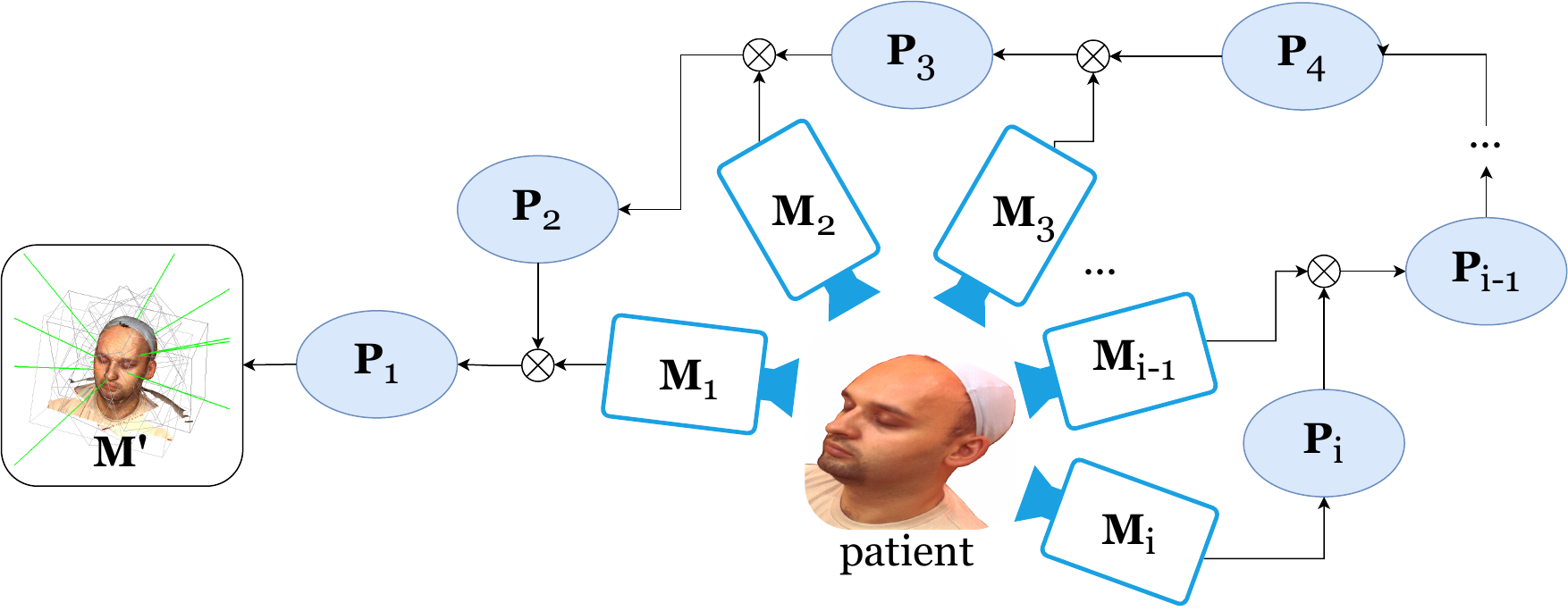}
    \caption{A group formed by the registered partial scans of the head and a resulting model}
    \label{fig:scanning_process}
\end{figure}

In a transformation tree, all elements of a group share the same transformation path. This principle applies, for example, to the registration of partial 3D face scans from different viewpoints, which can be performed pairwise, forming progressively expanding groups. Each newly registered scan is aligned with the previous one, forming a group. This group is then transformed using the preceding registration, progressively expanding until all scans are iteratively registered back to the first scan.
It is explained in Figure~\ref{fig:scanning_process} and can be represented with the following equation: 

\begin{equation}\resizebox{\linewidth}{!}{\begin{math}
\begin{array}{l}
\Big[\mathbf{M'}_{1}\quad\hdots\quad\mathbf{M'}_{i}\Big] = \\
\multicolumn{1}{r}{=\mathbf{P}_{1}\cdot\Bigg[
\mathbf{M}_{1}\quad\mathbf{P}_2\cdot\bigg[
\mathbf{M}_{2}\quad\mathbf{P}_3\cdot\Big[
\mathbf{M}_3\quad \mathbf{P}_4\cdot \big[
\hdots \quad \mathbf{P}_{i-1}\cdot \left[ \mathbf{M}_{i-1}\quad\mathbf{P}_{i}\cdot \mathbf{M}_{i} 
\right]
\big]
\Big]
\bigg]
\Bigg]}
\end{array}
\end{math}}\end{equation}

The transformation tree can be used to document the registration of  partial scans of gypsum dental models obtained from 3D scanning with a rotary table. 
The plaster dental models are models of the patient's teeth. They are created by taking impressions from both dental arches. Their mutual position is fixed by a bite bar. Scanning of each of the models on the rotary table is carried out by performing  scans correlated with the rotary table's movement. To obtain a watertight model, it is necessary to perform two or more series of scans for different positions of the plaster model on the rotary table.

Each part of the plaster model is scanned multiple times at different positions on the rotary table, producing several scan series. Registration occurs within each series, then between series, separately for the maxilla, mandible, and their occlusion. The transformation tree organizes registrations into three nested groups: registration within groups 
and then registration of groups between each other to get models of the jaw, mandible, and occlusion, and then registration of models of the maxilla and mandible to occlusion. 

\begin{figure}
    \centering
    \includegraphics[width=\linewidth]{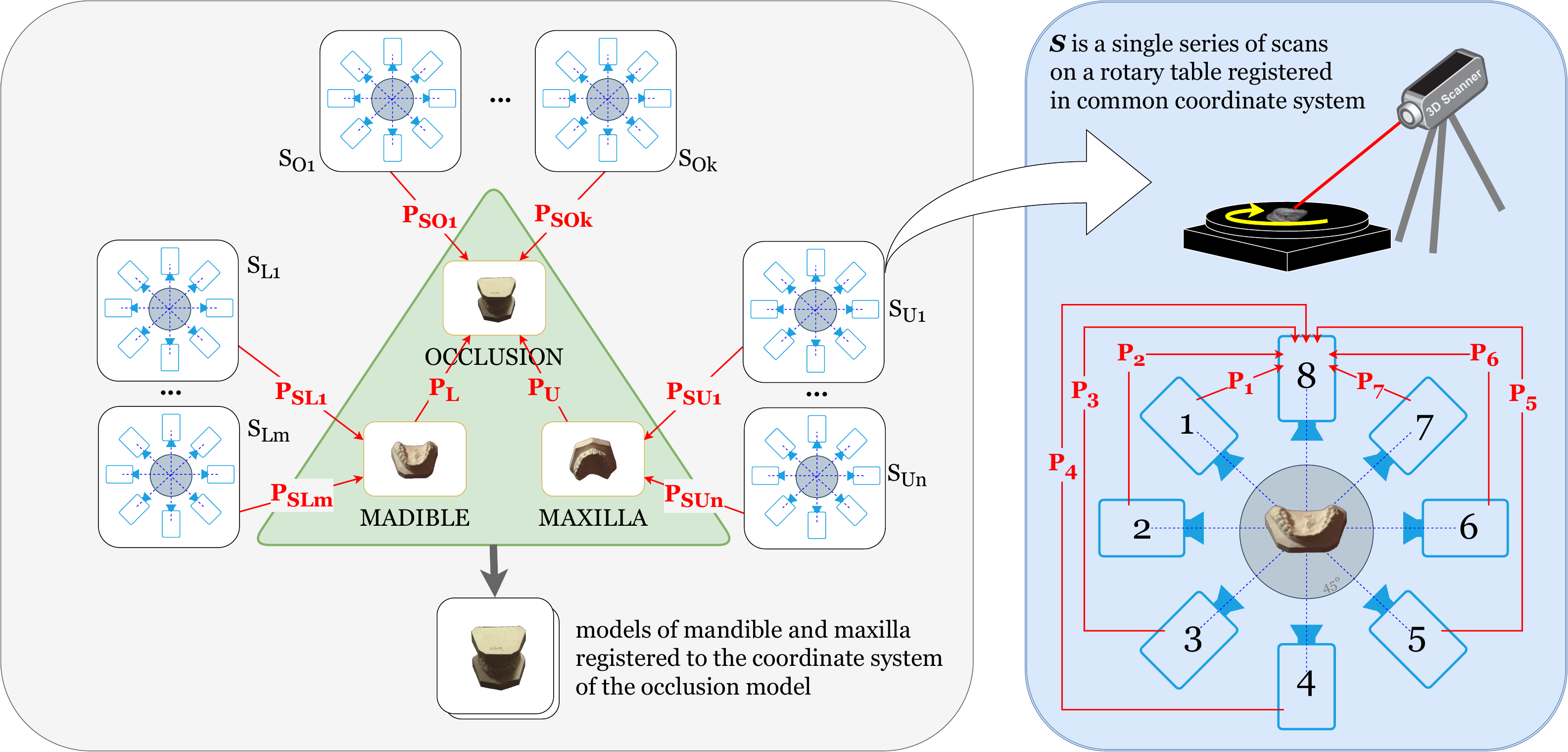}
    \caption{A diagram of successive registrations of partial 3D scans of plaster dental models.}
    \label{fig:stolik}
\end{figure}

Registration of scans obtained from scanning with a rotary table at a given angle of rotation comes down to finding the axis of rotation around which the table rotates. Although the article skips over aspects of registration algorithms, it should be mentioned that these specific assumptions impose limitations on the registration process. Once the intra-group scans have been matched, the groups should be registered to each other.  Since the transformation affecting a group affects the individual elements in the same way, as long as the single  elements of both groups can be matched with sufficient accuracy, the registration of these groups is automatically found.
A diagram of successive registrations of 3D scans of dental gypsum models scanned with a rotary table is shown in Figure~\ref{fig:stolik}.


\subsection { Documentation of motion}
\begin{figure}
    \centering
    \includegraphics[width=\linewidth]{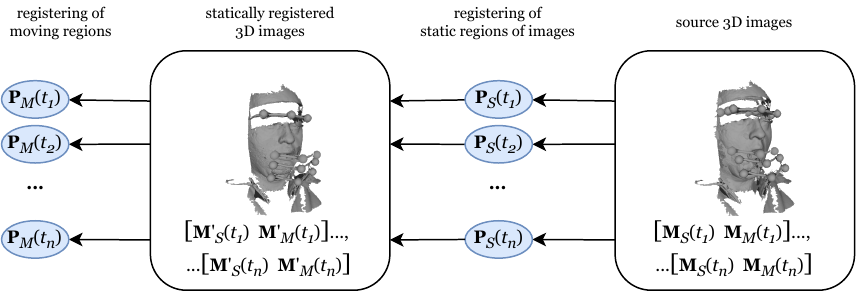}
    \caption{Movement acquisition scheme. All images in the captured sequence are first segmented to static ($\mathbf{M}_S$) and moving part ($\mathbf{M}_M$). The static part is used to find the correspondence $\mathbf{P}_{S}(t_i)$ between the images. Then, the correspondence of the moving parts is found in order to determine the transformation matrices $\mathbf{P}_{M}(t_i)$.}
    \label{fig:movement1}
\end{figure}

\begin{figure}
    \centering
    \includegraphics[width=\linewidth]{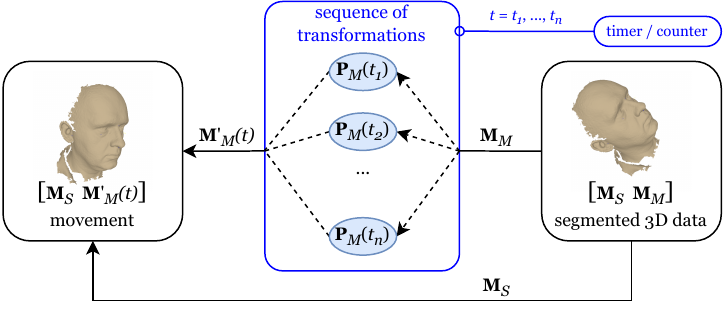}
    \caption{Scheme of visualisation of movement. $\mathbf{M}_S$ is a static part of 3D image and $\mathbf{M}_M$ is a moving part. The matrices determined in the motion capture process are sequentially used to position the moving image fragment.}
    \label{fig:animaion1}
\end{figure}
In addition to the analysis of the morphology of the stomatognathic system, the temporomandibular joint, its structure and its work remains in the field of interest of orthodontics.
Not all the components that constitute the temporomandibular joint can be imaged with the imaging described in the article. (i.e. the articular disc). However, an indirect examination is used, which allows us to assess the  function of this joint on the basis of movement. Movement in the stomatognathic system can be described as a change in the position of the mandible relative to the mandible.
Defining motion requires additional segmentation --- distinguishing  stable part from those subject to motion.

The movement can be treated as an object, such as a transformation, can be a group with other objects, and be subject to transformations that register a coordinate system in which the movement occurs to the coordinate system of other objects of the tree. The movement itself only affects the descendants of this object.

\begin{equation}
\Big[\mathbf{M'}_{S}\quad\mathbf{M'}_{M}\left(t\right)\Big] = \mathcal{P}\cdot
\Big[\mathbf{M}_{S}\quad\mathbf{P}_M\left(t\right)\cdot\mathbf{M}_{M}\Big]
\end{equation}
where $\mathbf{P}_M\left(t\right)$ is the movement represented as transformations in $t$ moment of time, $\mathbf{M}_{S}$, $\mathbf{M}_{M}$ --- are representations of stable and moving parts.

The function that defines the motion can be obtained from different systems using different measurement methods, like \textit{Zebris JMA--Optic}, \textit{axiographs}, and \textit{3D dynamic scanners} or other vision systems \cite{Bydgoszcz2019,zebris,Jakubowska2023,tomaka:itib16}.

\section{Discussion}\label{sec:discussion}

The main advantage of using transformation trees to describe patient data is that they introduce order by grouping objects subjected to the same transformations. Storing transformation coefficients also eliminates the need to save images in multiple coordinate systems, reducing the memory required for storing the entire dataset.

While visualizing the final model requires converting it into the coordinate system associated with the user's viewpoint, applying the transformation tree simplifies this process. The necessary calculations involve only a single multiplication with the latest overall transformation matrix.

The transformation tree links all objects that constitute the patient's image dataset. This means that by recalculating transformations from a given image to the root and then from the root to another image, the first image can be represented in the coordinate system of the second, forming a group with it. 

\begin{figure}
    \centering
    \includegraphics[width=\linewidth]{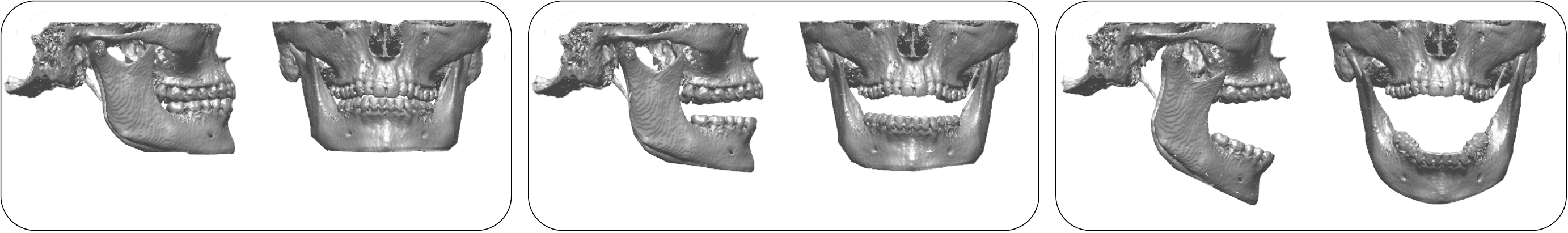}
    \caption{Simulation of mandible movement.}
    \label{fig:klapanie}
\end{figure}

This method can be also used to apply motion data to a group of mandibular images. It can also be integrated with CBCT data for further analysis. Such integration allows for the simulation of mandibular movement (Figure~\ref{fig:klapanie}) and enables dynamic analysis of changes in the mandibular condyles, TMJ, and occlusion.

\begin{figure}
    \centering
    \includegraphics[width=\linewidth]{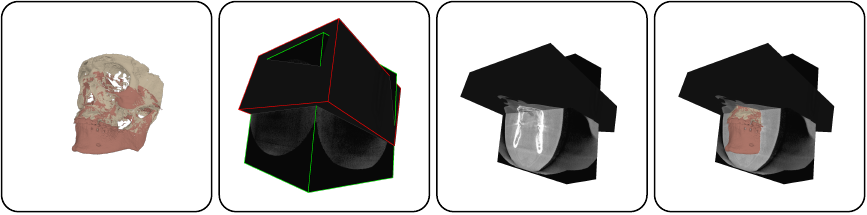}
    \caption{Indirect registration of volumetric images using the alignment of 3D bone reconstructions. The resulting transformation is applied to entire groups of images.}
    \label{fig:volum2}
\end{figure}

This also means that when a transformation to any element of the tree can be found, the new image can be registered to the overall patient model. 

For example, if a transformation can be determined between two bone tissue reconstructions, this registration can be applied to the entire group of volumetric images and the corresponding segmentations (Figure~\ref{fig:volum2}).

Using a tree structure to record transformations leading to image registration and patient model creation also has certain limitations. According to tree structure theory, if there is more than one path between two nodes, the structure is no longer a tree but a graph. 

Such a situation may occur when registering two types of images belonging to different groups (e.g., separate registrations for bone and soft tissues). These registrations may result in slightly different transformations due to possible soft tissue deformations or varying segmentation thresholds. This issue highlights the need for further research on selecting an optimal structure for representing such transformations, including assigning weights that account for the quality of the obtained registrations.

\section{Conclusions}\label{sec:conclusions}

The key conclusion is that the transformation tree establishes interrelationships between all images by organizing them and documenting the sequence of transformations from the source image to the final patient-defined coordinate system. This method records all operations performed on the images, ensuring reproducibility and enabling the indirect alignment of images that lack a common reference structure. By propagating transformations across the tree, new images can be consistently positioned within the patient model, even when direct registration is not feasible.

The proposed approach is versatile enough to be applied in other fields involving computer graphics and multimodal imaging. Its ability to manage hierarchical transformations efficiently makes it a promising tool for applications requiring precise spatial alignment, such as biomechanical simulations or forensic image analysis. While the transformation tree provides a systematic framework for organizing transformations, its limitations should also be acknowledged. In cases where multiple registration paths exist between two nodes, the tree structure transitions into a graph, introducing challenges related to transformation consistency. Future research should focus on extending this model to incorporate confidence weights for transformation paths and evaluating its robustness in scenarios with complex deformations.





%
%


\section*{Acknowledgements}

The authors would like to thank Krzysztof Domino, PhD, for his valuable advice and motivation, which supported the creation of this article.

\section*{Declaration of competing interest}
The authors have no competing interests to declare that are relevant to the content of this article.


\begin{thebibliography}{10}
\expandafter\ifx\csname url\endcsname\relax
  \def\url#1{\texttt{#1}}\fi
\expandafter\ifx\csname urlprefix\endcsname\relax\def\urlprefix{URL }\fi
\expandafter\ifx\csname href\endcsname\relax
  \def\href#1#2{#2} \def\path#1{#1}\fi

\bibitem{joda2014}
T.~Joda, G.~O. Gallucci, The virtual patient in dental medicine, Clinical Oral Implants Research 26~(6) (2015) 725--726.
\newblock \href {https://doi.org/10.1111/clr.12379} {\path{doi:10.1111/clr.12379}}.

\bibitem{marradi}
F.~Marradi, E.~Staderini, M.~A. Zimbalatti, A.~Rossi, C.~Grippaudo, P.~Gallenzi, How to obtain an orthodontic virtual patient through superimposition of three-dimensional data: A systematic review, Applied Sciences 10~(15) (2020).
\newblock \href {https://doi.org/10.3390/app10155354} {\path{doi:10.3390/app10155354}}.

\bibitem{seth}
I.~Seth, B.~Lim, P.~Y.~J. Lu, Y.~Xie, R.~Cuomo, S.~K.-H. Ng, W.~M. Rozen, F.~Sofiadellis, Digital twins use in plastic surgery: A systematic review, Journal of Clinical Medicine 13~(24) (2024).
\newblock \href {https://doi.org/10.3390/jcm13247861} {\path{doi:10.3390/jcm13247861}}.

\bibitem{LO2023100615}
L.-J. Lo, H.-H. Lin, Applications of three-dimensional imaging techniques in craniomaxillofacial surgery: A literature review, Biomedical Journal 46~(4) (2023) 100615.
\newblock \href {https://doi.org/10.1016/j.bj.2023.100615} {\path{doi:10.1016/j.bj.2023.100615}}.

\bibitem{Golub1973LeastSquares}
G.~H. Golub, C.~F.~V. Loan, An analysis of the total least squares problem, SIAM Journal on Numerical Analysis 10~(2) (1973) 241--256.
\newblock \href {https://doi.org/10.1137/0710024} {\path{doi:10.1137/0710024}}.

\bibitem{besl}
P.~Besl, N.~D. McKay, A method for registration of 3-d shapes, IEEE Transactions on Pattern Analysis and Machine Intelligence 14~(2) (1992) 239--256.
\newblock \href {https://doi.org/10.1109/34.121791} {\path{doi:10.1109/34.121791}}.

\bibitem{POJDA2025102093}
D.~Pojda, M.~Żarski, A.~A. Tomaka, L.~Luchowski, dpvision: Environment for multimodal images, SoftwareX 30 (2025) 102093.
\newblock \href {https://doi.org/10.1016/j.softx.2025.102093} {\path{doi:10.1016/j.softx.2025.102093}}.

\bibitem{dpvision}
D.~Pojda, dp{V}ision (data processing for vision), opensource software repository, accessed: 2025-01-10 (2024).
\newblock \href {https://doi.org/10.5281/zenodo.13944334} {\path{doi:10.5281/zenodo.13944334}}.

\bibitem{Ritter2011MeVisLab}
F.~Ritter, T.~Boskamp, A.~Homeyer, H.~Laue, M.~Schwier, F.~Link, H.-O. Peitgen, Medical image analysis: A visual approach, IEEE Pulse 2~(6) (2011) 60--70.
\newblock \href {https://doi.org/10.1109/MPUL.2011.942929} {\path{doi:10.1109/MPUL.2011.942929}}.

\bibitem{McCormick2014ITK}
M.~McCormick, S.~Liu, J.~Ibanez, S.~Jomier, L.~Marion, Itk: Enabling reproducible research and open science, Frontiers in Neuroinformatics 8 (2014) 13.
\newblock \href {https://doi.org/10.3389/fninf.2014.00013} {\path{doi:10.3389/fninf.2014.00013}}.

\bibitem{Klein2010Elastix}
S.~Klein, M.~Staring, K.~Murphy, M.~A. Viergever, J.~P.~W. Pluim, elastix: A toolbox for intensity-based medical image registration, IEEE Transactions on Medical Imaging 29~(1) (2010) 196--205.
\newblock \href {https://doi.org/10.1109/TMI.2009.2035616} {\path{doi:10.1109/TMI.2009.2035616}}.

\bibitem{meshlab2008}
P.~Cignoni, M.~Callieri, M.~Corsini, M.~Dellepiane, F.~Ganovelli, G.~Ranzuglia, {MeshLab: an Open-Source Mesh Processing Tool}, in: V.~Scarano, R.~D. Chiara, U.~Erra (Eds.), Eurographics Italian Chapter Conference, The Eurographics Association, 2008.
\newblock \href {https://doi.org/10.2312/LocalChapterEvents/ItalChap/ItalianChapConf2008/129-136} {\path{doi:10.2312/LocalChapterEvents/ItalChap/ItalianChapConf2008/129-136}}.

\bibitem{girardeau2014cloudcompare}
D.~Girardeau-Montaut, Cloud{C}ompare, a 3d point cloud and mesh processing free software, EDF R\&D, Telecom ParisTech (2014).

\bibitem{Tustison2021ANTsX}
N.~J. Tustison, P.~A. Cook, A.~J. Holbrook, H.~J. Johnson, J.~Muschelli, G.~A. Devenyi, J.~T. Duda, S.~R. Das, N.~C. Cullen, D.~L. Gillen, M.~A. Yassa, J.~R. Stone, J.~C. Gee, B.~B. Avants, The antsx ecosystem for quantitative biological and medical imaging, Scientific Reports 11 (2021) 1--12.
\newblock \href {https://doi.org/10.1038/s41598-021-87564-6} {\path{doi:10.1038/s41598-021-87564-6}}.

\bibitem{Fedorov2012Slicer}
A.~Fedorov, R.~Kikinis, S.~Pieper, J.-C. Fillion-Robin, S.~Pujol, J.~Finet, C.~Pinter, A.~Lasso, J.~Miller, S.~Aylward, et~al., 3d slicer as an image computing platform for the quantitative imaging network, Magnetic Resonance Imaging 30~(9) (2012) 1323--1341.
\newblock \href {https://doi.org/10.1016/j.mri.2012.05.001} {\path{doi:10.1016/j.mri.2012.05.001}}.

\bibitem{Hunter2001XML}
J.~Hunter, The application of metadata standards to video indexing, IEEE Multimedia 8~(4) (2001) 30--37.
\newblock \href {https://doi.org/10.1109/93.959093} {\path{doi:10.1109/93.959093}}.

\bibitem{Crockford2006JSON}
D.~Crockford, The application/json media type for javascript object notation (json), RFC Editor 4627 (2006) 1--10.
\newblock \href {https://doi.org/10.17487/RFC4627} {\path{doi:10.17487/RFC4627}}.

\bibitem{Khronos2017glTF}
K.~Group, \href{https://www.khronos.org/gltf/}{gltf - the gl transmission format}, Khronos Group Technical Report (2017).
\newline\urlprefix\url{https://www.khronos.org/gltf/}

\bibitem{Wavefront1992OBJ}
W.~Technologies, \href{https://www.fileformat.info/format/wavefrontobj/}{The obj file format specification}, Wavefront Technologies Documentation (1992).
\newline\urlprefix\url{https://www.fileformat.info/format/wavefrontobj/}

\bibitem{Attene2018STL}
M.~Attene, As-exact-as-possible repair of unprintable stl files, Rapid Prototyping Journal 24~(2) (2018) 348--357.
\newblock \href {https://doi.org/10.1108/RPJ-11-2016-0185} {\path{doi:10.1108/RPJ-11-2016-0185}}.

\bibitem{Web3D2010VRML}
W.~Consortium, \href{https://www.web3d.org/x3d/}{X3d: Extensible 3d graphics for web authors}, Web3D Consortium Technical Report (2010).
\newline\urlprefix\url{https://www.web3d.org/x3d/}

\bibitem{Sharp2008DICOM}
G.~C. Sharp, K.~K. Lee, D.~K. Wehe, S.~Pieper, Dicom spatial registration: Framework and applications, Medical Physics 35~(11) (2008) 4978--4987.
\newblock \href {https://doi.org/10.1118/1.2990772} {\path{doi:10.1118/1.2990772}}.

\bibitem{McCormick2014DICOM}
M.~McCormick, K.~Wang, A.~Lasso, G.~Sharp, S.~Pieper, \href{http://hdl.handle.net/10380/3468}{Dicom spatial transform io in the insight toolkit}, Insight Journal 1 (2014) 1--8.
\newline\urlprefix\url{http://hdl.handle.net/10380/3468}

\bibitem{Hess2010BlenderFoundations}
R.~Hess, Blender Foundations: The Essential Guide to Learning Blender 2.5, Routledge, New York, 2010.
\newblock \href {https://doi.org/10.4324/9780240814315} {\path{doi:10.4324/9780240814315}}.

\bibitem{BlenderManual}
{Blender Online Community}, \href{https://www.blender.org}{Blender - a 3D modelling and rendering package}, Blender Foundation, Blender Institute, Amsterdam (2025).
\newline\urlprefix\url{https://www.blender.org}

\bibitem{Adeniji2024Unity}
I.~Adeniji, M.~Casarona, L.~Bielory, L.~Bancairen, M.~Menzel, N.~Perigo, C.~Blackmon, M.~G. Niepielko, J.~Insley, D.~Joiner, Using unity for scientific visualization as a course-based undergraduate research experience, Journal of Computational Science Education 15~(1) (2024) 35--40.
\newblock \href {https://doi.org/10.22369/issn.2153-4136/15/1/7} {\path{doi:10.22369/issn.2153-4136/15/1/7}}.

\bibitem{Kruger2024UnrealEngine}
M.~Krüger, D.~Gilbert, T.~W. Kuhlen, T.~Gerrits, Game engines for immersive visualization: Using unreal engine beyond entertainment, Presence: Virtual and Augmented Reality 33~(4) (2024) 11--13.
\newblock \href {https://doi.org/10.1162/pres\_a\_00416} {\path{doi:10.1162/pres\_a\_00416}}.

\bibitem{Bydgoszcz2019}
A.~A. Tomaka, L.~Luchowski, D.~Pojda, M.~Tarnawski, K.~Domino, \href{http://arxiv.org/abs/1911.08854}{The dynamics of the stomatognathic system from 4d multimodal data}, in: Gadomski, {A}. (ed.) {M}ultiscale {L}ocomotion: {I}ts Active-Matter Addressing Physical Principles, UTP University of Science \& Technology, Bydgoszcz, 2019, pp. 37--53.
\newline\urlprefix\url{http://arxiv.org/abs/1911.08854}

\bibitem{tomaka:2007}
A.~Tomaka, M.~Tarnawski, L.~Luchowski, B.~Lisniewska-Machorowska, Digital dental models and 3d patient photographs registration for orthodontic documentation and diagnostic purposes, in: M.~Kurzynski, E.~Puchala, M.~Wozniak, A.~Zolnierek (Eds.), Computer Recognition Systems 2, Springer Berlin Heidelberg, Berlin, Heidelberg, 2007, pp. 645--652.

\bibitem{solaberrieta2}
E.~Solaberrieta, R.~Mínguez, L.~Barrenetxea, O.~Etxaniz, Direct transfer of the position of digitized casts to a virtual articulator, The Journal of Prosthetic Dentistry 109~(6) (2013) 411--414.
\newblock \href {https://doi.org/10.1016/S0022-3913(13)60330-3} {\path{doi:10.1016/S0022-3913(13)60330-3}}.

\bibitem{solaberrieta3}
E.~Solaberrieta, A.~Garmendia, R.~Minguez, A.~Brizuela, G.~Pradies, Virtual facebow technique, The Journal of Prosthetic Dentistry 114~(6) (2015) 751--755.
\newblock \href {https://doi.org/10.1016/j.prosdent.2015.06.012} {\path{doi:10.1016/j.prosdent.2015.06.012}}.

\bibitem{Lama}
W.~Y. Lam, R.~T. Hsung, W.~W. Choi, H.~W. Luk, L.~Y. Cheng, E.~H. Pow, A clinical technique for virtual articulator mounting with natural head position by using calibrated stereophotogrammetry, The Journal of Prosthetic Dentistry 119~(6) (2018) 902--908.
\newblock \href {https://doi.org/10.1016/j.prosdent.2017.07.026} {\path{doi:10.1016/j.prosdent.2017.07.026}}.

\bibitem{zebris}
{Zebris~Medical}, Zebris {JMA}-{O}ptic {S}ystem. {L}ist of {D}ental {P}ublications, accessed: 2025-01-10. URL: https://www.zebris.de/fileadmin/Editoren/zebris-PDF/zebris-Literatur-PDF/Publications-Dental.pdf (2025).

\bibitem{Jakubowska2023}
S.~Jakubowska, M.~P. Szerszeń, J.~Kostrzewa-Janicka, Jaw motion tracking systems – literature review, Prosthodontics 73~(1) (2023) 18--28.
\newblock \href {https://doi.org/10.5114/ps/162663} {\path{doi:10.5114/ps/162663}}.

\bibitem{tomaka:itib16}
A.~A. Tomaka, M.~Tarnawski, D.~Pojda, Multimodal image registration for mandible motion tracking, in: Piętka {E}., {B}adura {P}., {K}awa {J}., {W}ięcławek {W}. (eds) {I}nformation {T}echnologies in {M}edicine. {ITiB} 2016. {A}dvances in {I}ntelligent {S}ystems and {C}omputing, vol 471, Springer, Cham, 2016, pp. 179--191.
\newblock \href {https://doi.org/10.1007/978-3-319-39796-2\_15} {\path{doi:10.1007/978-3-319-39796-2\_15}}.

\end{thebibliography}

\end{document}